# Leveraging Text Guidance for Enhancing Demographic Fairness in Gender Classification

## Anoop Krishnan

**Correspondence**
Anoop Krishnan
Email: anoopk682@gmail.com

## Abstract

In the quest for fairness in artificial intelligence, novel approaches to enhance it in facial image‑based gender classification algorithms using text‑guided methodologies are presented. The core methodology involves leveraging semantic information from image captions during model train‑ ing to improve generalization capabilities. Two key strategies are presented: Image‑Text Matching (ITM) guidance and Image‑Text fusion. ITM guidance trains the model to discern fine‑grained alignments between images and texts to obtain enhanced multimodal representations. Image‑text fusion combines both modalities into comprehensive representations for improved fairness. Ex‑ tensive experiments conducted on benchmark datasets demonstrate these approaches effectively mitigate bias and improve accuracy across gender‑racial groups compared to existing methods. Ad‑ ditionally, the unique integration of textual guidance underscores an interpretable and intuitive training paradigm for computer vision systems. By scrutinizing the extent to which semantic in‑ formation reduces disparities, this research offers valuable insights into cultivating more equitable facial analysis algorithms. The proposed methodologies contribute to addressing the pivotal chal‑ lenge of demographic bias in gender classification from facial images, furthermore, this technique operates in the absence of demographic labels and is application‑agnostic.

## 1 | INTRODUCTION

Artificial intelligence (AI) systems are increasingly utilized for high‑ stake decision‑making applications including border control, criminal justice, and healthcare Mittermaier et al. (2023), Carlos‑Roca et al. (2018), Movva (2021). Consequently, it is crucial that these systems do not exhibit bias. However, recent research has revealed significant accu‑ racy disparities indicating bias in AI‑based systems across demographic groups such as gender and race Krishnan et al. (2020), Albiero et al. (2020). The decision of a biased system is skewed toward a particular demographic sub‑group.

Facial analysis‑based systems are at the center stage of this dis‑ cussion Krishnan et al. (2020), Albiero et al. (2020). Automated facial analysis includes face detection, recognition, and attribute classification (including gender‑, race‑, age classification, and BMI prediction) Siddiqui et al. (2022), Nadimpalli and Rattani (2022), Levi and Hassner (2015), Al‑ madan et al. (2020), Zhang et al. (2017b), Masood et al. (2018). Several studies have analyzed the fairness of this facial‑analysis based algo‑ rithms; and confirmed performance disparities for females and people of color Grother et al. (2011), Klare et al. (2012), Best‑Rowden and Jain (2018), Abdurrahim et al. (2018), Raji and Buolamwini (2019), Vera‑ Rodríguez et al. (2019), Albiero et al. (2020), Krishnan et al. (2020), Muthukumar (2019). These facial analysis‑based algorithms are being deployed in various applications such as surveillance, retail, healthcare, and education. Due to their widespread use, developing accurate and unbiased facial analysis algorithms has become an urgent need for the deployment of fair and trustworthy systems.

Among various facial attributes, demographic gender attribute has garnered significant attention Ricanek and Tesafaye (2006), Levi and Hassner (2015), Theobald et al. (2017), Buolamwini and Gebru (2018), Gbekevi et al. (2023), Roxo and Proença (2021), being integrated into various applications such as image retrieval, surveillance, and human-computer interaction. Automated gender classification is often fused with primary biometric modalities, like facial and ocular recognition, to enhance user identification accuracy. Notably, industry leaders like Amazon Rekognition Rekognition (2022), DeepVision AI Vision (2022), FaceX FaceX (2022), and Microsoft Azure Cognitive Services Services (2022) have incorporated automated face-based gender classification into their commercial software development kits (SDKs), highlighting the widespread adoption and practical significance of this technology.

---

† The word systems and algorithms are used interchangeably in this paper.

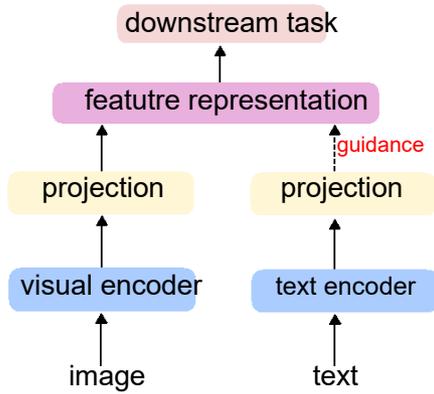

**FIGURE 1** Overview of our proposed text-guided visual attribute classifier .

Several studies confirm the bias of face-based gender classification algorithms across gender-racial groups Ricanek and Tesafaye (2006), Levi and Hassner (2015), Theobald et al. (2017), Buolamwini and Gebru (2018), Gbekevi et al. (2023), Roxo and Proença (2021). In response, nu- merous bias mitigation approaches have been proposed by the vision community to address gender classification bias. These approaches en- compass a variety of techniques, including consistency-regularization based multiple views Krishnan and Rattani (2023), attention mecha- nism Majumdar et al. (2021), adversarial debiasing Zhang et al. (2018), Chuang and Mroueh (2021), GAN-based oversampling Zietlow et al. (2022a), Ramachandran and Rattani (2022), multi-task classification Das et al. (2018), network pruning Lin et al. (2022) and one using human- machine partnership Nair and Rattani (2025). Since most of the research on this subject adheres to the idea that "gender" is binary, we also stick to it for the sake of fair comparison. While conducting this study, we do not intend to belittle people who disagree with this view. Moreover, gender labels used in this study were sourced from the provided anno- tation with the publicly available facial datasets used for both training and validation.

Despite their efficacy, current bias mitigation techniques for gender classification often rely on demographically annotated training datasets or involve computationally intensive processes, limiting their applicabil- ity and generalizability. Additionally, these methods may compromise overall classification accuracy by enhancing fairness, thereby exhibiting *Pareto inefficiency* Zhang et al. (2018).

Recently, studies in Krishnan and Rattani (2023), Li and Vasconcelos (2019) proposed promising Pareto-efficient Zietlow et al. (2022b) bias mitigation techniques that **optimizes the learned feature representa- tion** for each demographic sub-group using adaptive data resampling Li and Vasconcelos (2019) and multi-view based consistency regulariza- tion Krishnan and Rattani (2023). These techniques support the notion that better feature representation contributes to the model's gener- alizability and variance reduction in underrepresented groups. The ef- fectiveness of these mitigation techniques has been demonstrated in action recognition, and gender classification tasks.

Concurrently, there is a noticeable surge in research endeavors integrating natural language supervision for image feature representa- tion Radford et al. (2021), Jia et al. (2021), Singh et al. (2022a) for different tasks such as image classification, retrieval, and so on. The ap- proach uses textual descriptions of the images as a guide during the model's training process. It draws inspiration from the human learning process, where traditional pedagogical methods involve educators elu- cidating image characteristics through natural language explanations. However, these methods are not specifically designed to mitigate bias in image representation, especially for biometric applications like facial attribute classification.

Combining the paradigms of learning optimizing feature representa- tions across demographic subgroups and integrating natural language supervision, this paper aims to leverage the guidance from rich semantic context embedded within textual descriptions to enhance the demo- graphic fairness of a vision classifier. Figure 1 illustrates the general overview of this approach of using textual descriptions as a guiding mechanism during training for learning the better feature represen- tation for the vision classifier for any downstream classification task. Consequently, our study investigates the extent to which semantic information integration from textual descriptions minimizes representa- tion biases and disparities across diverse racial-gender groups. To this front, we have proposed two novel solutions for integrating textual guid- ance during the training stage of the vision classifier as a case study on gender classification.

## 1.1 | Our Contribution

In summary, the main contributions of this work are as follows:

- A first-of-its-kind study based on leveraging the text modality/guid- ance for enhanced fairness of the vision classifiers as a case study on gender classification.

- Proposed two novel techniques, firstly an image classification ap- proach guided by an image-text matching framework, where the alignment between visual and linguistic representations is utilized to enhance image classification performance.

- Secondly, an image classification method fuses the representations of both the image and corresponding text modalities, creating a com- bined representation incorporating information from both modali- ties.

- Extensive evaluation of the proposed bias mitigation strategy in the intra- and cross-dataset evaluation scenario. To this front, CelebA

dataset Liu et al. (2015) for training, and FairFace Kärkkäinen and Joo (2019), DiveFace Morales et al. (2021), and UTKFace Zhang et al. (2017a) datasets were used for the evaluations.

- Cross‐comparison with the SOTA bias mitigation techniques for gender classification.
- Multiple ablation studies are performed to identify the optimal textual information using prompt engineering and human analysis for each image, aiming to enhance fairness in the vision classification system. These studies contribute nuanced insights into the role of textual guidance in mitigating biases, providing valuable guidance for refining and optimizing the proposed methodology.

This paper is organized as follows: Section 2 reviews relevant literature on text‐guided vision classification and evaluation and mitigation of bias in gender classification algorithms. Section 3 presents the proposed bias mitigation strategy. Section 4 describes the datasets, implementa‐ tion details, and evaluation metrics used in this study. Additionally, it discusses the results, ablation study findings, and potential future re‐ search directions. Finally, Section 5 summarizes the key findings and concluding remarks.

## 2 | RELATED WORK

### 2.1 | Text‐guided vision classification

Integrating text guidance into vision classification marks a significant advancement in computer vision. Zhang et al. Zhang et al. (2021) proposed TandemNet which is based on learning interactions between visual and semantic text information using a dual attention mechanism for learning useful features. Their work when applied to the medical domain contributes to enhanced performance and interpretability by highlighting discriminative and informative image regions and the text description.

The significance of text‐guided vision models is further emphasized by the findings of CLIP (Contrastive Language–Image Pre‐training) Radford et al. (2021). This model learns a multi‐modal embedding space through joint training of image and text encoders, enhancing robustness for various vision‐based tasks. The authors substantiated that zero‐shot CLIP models exhibit significantly greater robustness compared to equivalently accurate supervised ImageNet‐based vision models. Subsequent models like ALIGN Jia et al. (2021), FLAVA Singh et al. (2022a), and Florence Yuan et al. (2021) further refined the approach of using natural language supervision for image representation learning. Notably, these models have demonstrated success across diverse tasks such as classification, detection, and segmentation, highlighting the versatility and effectiveness of text‐guided approaches in addressing various computer vision problems across different domains. However, these models were not specifically developed to reduce demographic biases in their outputs. Further, Zhang et al. Zhang et al. (2023) proposed the Connect Image and Text Embeddings (CITE) method to enhance pathological image classification. The key idea is to inject text insights from large lan‐ guage models into the image classification task, thereby adapting the foundation model toward better pathological image understanding. Im‐ portantly, CITE introduces only a small number of trainable parameters to the pre‐trained foundation model, making it an efficient approach for leveraging multimodal information.

## 2.2 Bias Evaluation and Mitigation of Gender Classification

The foundational literature underscores the inherent biases present in gender classification algorithms, particularly concerning specific racial and gender demographics, as documented by Buolamwini et al. Buo‐ lamwini and Gebru (2018), Raji and Buolamwini Raji and Buolamwini (2019), Muthukumar Muthukumar (2019), Balakrishnan et al. Balakrish‐ nan et al. (2020), and Krishnan et al. Krishnan et al. (2020).

Muthukumar Muthukumar (2019) and Balakrishnan et al. Balakr‐ ishnan et al. (2020) suggested that factors such as age, hair length, and facial hair contribute significantly to performance differentials, particularly for women and individuals with darker skin tones, when evaluations on the PPB dataset. Krishnan et al. Krishnan et al. (2020) extended this analysis by evaluating various CNN architectures (ResNet‐ 50, Inception‐V4, VGG‐ 16/19, and VGGFace) for gender classification across diverse gender‐racial groups on the UTKFace and FairFace datasets. The findings highlight the impact of the CNN architectural dif‐ ferences on uneven accuracy rates and attribute black females' high misclassification error rate to the morphological similarity with black males.

Numerous strategies have been proposed to address the inherent bi‐ ases in gender classification algorithms. Teru and Chakraborty Teru and Chakraborty (2019) introduced an adversarially learned encoder to ob‐ tain ethnicity‐invariant representations for gender classification, when evaluated on the UTKFace dataset. Das et al. Das et al. (2018) proposed a Multi‐Task Convolution Neural Network (MTCNN) capable of jointly classifying gender, age, and ethnicity, while concurrently minimizing the impact of protected attributes. The efficacy of this model was evaluated on both the UTKFace and BEFA datasets.

Majumdar et al. Majumdar et al. (2021) proposed an attention‐aware debiasing method, employing an attention module to emphasize features pertinent to the main classification task while suppressing those associated with sensitive attributes. Experimental validation was con‐ ducted on the UTKFace and Morph datasets.

Ramachandran and Rattani Ramachandran and Rattani (2022), along with Ramaswamy et al. Ramaswamy et al. (2021), proposed solutions utilizing generative views obtained through GAN‐based latent vec‐ tor editing complemented by structured learning for enhanced representation learning and mitigating gender classification bias. Krishnan and Rattani Krishnan and Rattani (2023) introduced a consistency regularization‐based model utilizing image‐level and feature‐level aug‐ mentation to alleviate bias, with demonstrated effectiveness on the FairFace, UTKFace, and DiveFace datasets.

Park et al. Park et al. (2022) proposed the Fair Supervised Contrastive Loss (FSCL) designed to enforce similarity in the feature representation between samples for the same class against different classes, contribut- ing to fair representation learning. The effectiveness of this approach was substantiated through empirical evaluation on CelebA and UTKFace datasets.

# 3 | PROPOSED APPROACH

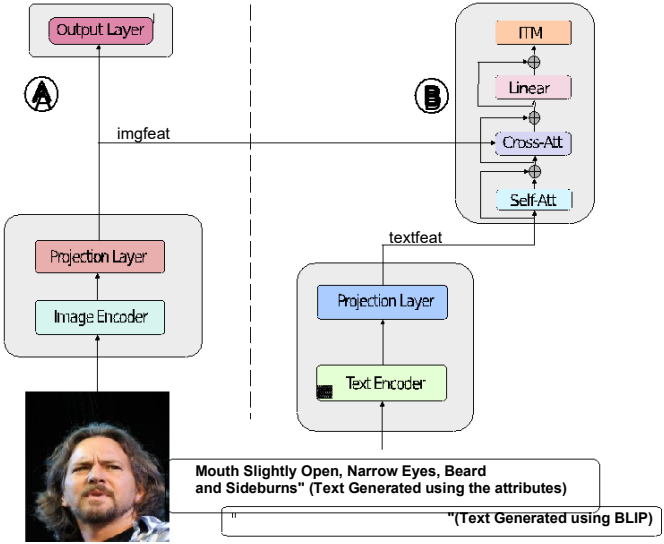

**F I G U R E 2** Proposed image classification guided by image-text matching.

In this section, novel approaches for text-guided image classification in the context of gender classification from facial images, with a specific emphasis on *image-only operation during test time*, and further improving fairness compared to other state-of-the-art facial image-based gender classification methods are detailed.

## 3.1 | Image-Text Matching (ITM) for guid-ance

In the domain of leveraging text-based guidance for gender classifi- cation, we propose an innovative model designed on the image-text matching principle to acquire a multimodal representation (MMR). The primary objective is to discern nuanced alignment between textual and visual information, where the model is tasked with determining whether a given text matches an accompanying image. This model functions as a guiding framework to enhance image classification tasks by incorporat- ing a fine-grained understanding of the intricate relationship between image and text.

The process can be explained using Figure 2, which takes an image and its accompanying text as input. Both the image and text are passed through their respective encoders (image encoder and text encoder) to extract features. These features are then projected through linear layers to obtain task-specific representations, making the model more adaptable and efficient for downstream classification tasks. The image and text representations are then combined in the Image-Text Match- ing (ITM) module, illustrated in Section **B** in Figure 2. This module consists of a multi-head self-attention layer followed by a multi-head cross-attention layer to combine the image and text features as shown in Equation 2. The combined representation is then passed through a linear layer, which helps adapt it for the image-text matching task by learning a projection. Finally, a binary output layer provides the prediction.

The combination of Sections **A** (image-only module) and **B** (ITM module) forms the comprehensive module designed for gender classifi- cation leveraging guidance from the image-text matching process.

$$Attention(Q, K, V) = softmax\left(\frac{QK^T}{\sqrt{d}}V\right) \quad (1)$$

$$MMR = Attention(imgfeat, textfeat, textfeat) + \\ Attention(textfeat, imgfeat, imgfeat) \quad (2)$$

Algorithm 1 provides step-by-step details of the aforementioned ITM method. The input $x_v$ and $x_t$ represent an image and the accompanying text and $y$ and $y_{match}$ represent the true label of the image and true label which indicates whether image-text matches (0/1 - does not match/- match). During the training stage, the algorithm takes an image $x_v$ and its corresponding text $x_t$ as input. The image $x_v$ is passed through an ImageEncoder to obtain its encoded representation, which is stored in $x_v$. Similarly, the text $x_t$ is passed through a TextEncoder to obtain its encoded representation, which is stored in $x_t$. The encoded image rep- resentation $x_v$ is then projected using a $Projection_v$ layer to obtain the image feature representation $imgfeat$. The encoded text representation $x_t$ is projected using a $Projection_t$ layer to obtain the text feature repre- sentation $textfeat$. The $imgfeat$ and $textfeat$ are then passed through the Image-Text Matching (ITM) module $f_{itm}$ to obtain the predicted match $y'_{match}$ between the image and text. The encoded image representation $x_v$ is also passed through a classification layer $f_{clf}$ to obtain the predicted classification output $y'$.

Two loss functions are calculated during training: $loss_1$ is the classi- fication loss between the true classification label $y$ and the predicted classification output $y'$. This is the primary loss for the gender classifi- cation task. $loss_2$ is the classification loss between the true match label $y_{match}$ and the predicted match $y'_{match}$ from the Image-Text Matching (ITM) module. This auxiliary loss aims to learn a multimodal represen- tation that captures fine-grained alignment between the image and its corresponding text.

The total training loss is the sum of these two losses, as shown in Equation 3:

$$TotalLoss = loss_1 + loss_2 \quad (3)$$

The ITM module is treated as a binary classification task, where a linear layer (ITM head) predicts whether an image-text pair is matched or unmatched based on their fused multimodal features. To create informa- tive negative examples, the image is captioned with its incorrect label during training.

During the testing stage, The algorithm takes a test image $x^t_v$ as in- put. The test image $x^t_v$ is passed through the Image Encoder to obtain its encoded representation, which is stored in $x_v$. The encoded test im- age representation $x_v$ is projected using $Projection_v$ to obtain the image feature representation $imgfeat$. The $imgfeat$ is passed through the classi- fication layer $f_{clf}$ to obtain the final classification output, which is stored in $output$.

---

**Algorithm 1** Pseudocode for ITM-Guided Algorithm

**Require:** $x_v$, $x_t$                    Training inputs: Images and Text
**Ensure:** $y$, $y_{match}$                 Outputs: Classification and Matching Scores

  **Training Stage:**

1: $f_{itm}$: *Image-Text Matching Module*, $f_{clf}$: *Classification Layer*
2: **for** each $(x_v, x_t)$ in *Images, Text* **do**
3:    $x_v \leftarrow$ ImageEncoder($x_v$)
4:    $x_t \leftarrow$ TextEncoder($x_t$)
5:    $imgfeat \leftarrow$ Projection$_v$($x_v$)
6:    $textfeat \leftarrow$ Projection$_v$($x_t$)
7:    $y'_{match} \leftarrow f_{itm}(imgfeat, textfeat)$
8:    $y' \leftarrow f_{clf}(imgfeat)$
9:    $loss_1 \leftarrow$ ClassificationLoss($y_{match}$, $y'_{match}$)
10:   $loss_2 \leftarrow$ ClassificationLoss($y$, $y'$)
11:   $loss \leftarrow loss_1 + loss_2$
12: **end for**

  **Testing Stage:**

**Require:** $x^t_v$                         Testing input: Images
13: **for** each $x^t_v$ in *Images* **do**
14:   $x_v \leftarrow$ ImageEncoder($x^t_v$)
15:   $imgfeat \leftarrow$ Projection$_v$($x_v$)
16:   $output \leftarrow f_{clf}(imgfeat)$
17: **end for**

---

## 3.2 | Image-Text fusion

As illustrated in Figure 3, this approach leverages information derived from image-text feature embedding through fusion techniques. This so- phisticated fusion process involves the amalgamation of both image and text information through self-attention, where the image and text em- beddings are concatenated as explained in Equation 5 to create a more comprehensive and discriminative feature representation.

The feature fusion process employs a three-layered architecture, comprising a multihead self-attention layer sandwiched by two linear layers, as depicted in Section 2 of C in Figure 2. The initial layer processes inputs with concatenated features, subsequently fed into the self-attention layer, as outlined in Equation 5. The final layer generates the fused feature, which is then input to the gender classification layer.

Algorithm 2 gives an overview of this method.

$$fusedfeat = imgfeat \ textfeat \qquad (4)$$

$$imagetextfeat = Attention(fusedfeat, \\ fusedfeat, fusedfeat) \qquad (5)$$

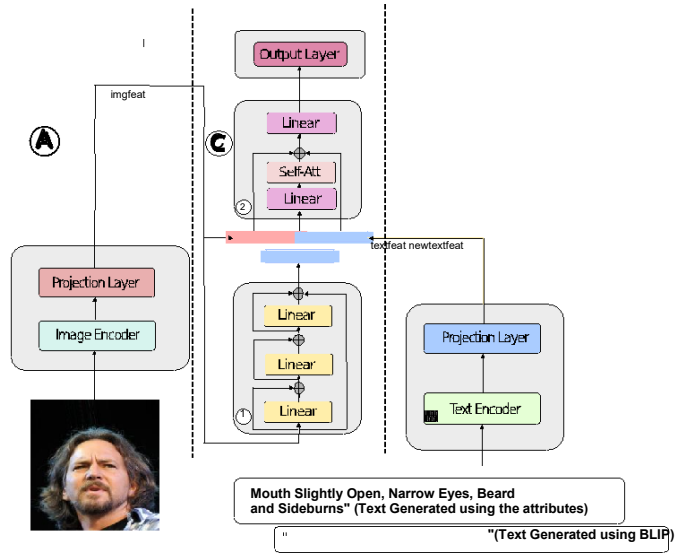

**FIGURE 3** Proposed image classification by image-text fusion.

However, during testing, we do not rely on explicit text descriptions. To generate text embeddings in the absence of captions, we utilized meta-learning techniques Biggs (2011), Bengio et al. (1991) to train a module that can produce corresponding textual embeddings from the image embeddings. This module optimizes the following objective to emulate textual embeddings:

$$\phi_c = arg_{\phi_c} \min E_{p(x_1, x_2)}(-\log(p(\hat{x}^2|x^1; \phi_c)) \qquad (6)$$

where $x_1$ is the image embedding, $x_2$ and $\hat{x}_2$ are textual embedding and generated textual embedding respectively by the module with parameters $\phi_c$.

The core idea is to perturb the latent visual feature space, enabling the image embeddings to approximate the text modality embeddings. We use a three-layer Multilayer Perceptron (MLP) with skip connections, shown in Section 1 of C in Figure 3, to generate text embeddings from image embeddings during testing. In essence, this meta-learned module allows us to synthesize textual representations from visual data when no actual text is available, facilitating the image‐text fusion process during inference.

Algorithm 2 provides step‐by‐step details of the aforementioned Image‐Text Fusion method. During the training stage, the algorithm takes an image $x_v$ and its corresponding text $x_t$ as input. The image $x_v$ is passed through an ImageEncoder to obtain its encoded representation, which is stored in $x_v$. Similarly, the text $x_t$ is passed through a TextEn‐ coder to obtain its encoded representation, which is stored in $x_t$. The encoded image representation $x_v$ is then projected using a $Projection_v$ layer to obtain the image feature representation $imgfeat$. The encoded text representation $x_t$ is projected using a $Projection_t$ layer to obtain the text feature representation $textfeat$. The $imgfeat$ is passed through the TextFeatGen module, which generates a new text feature representa‐ tion $newtextfeat$ from the image features. The $imgfeat$ and $textfeat$ are fused using the ImgTextFuse module to obtain the fused image‐text fea‐ ture representation $imagetextfeat$. Similarly, the $imgfeat$ and $newtextfeat$ are fused using the ImgTextFuse module to obtain the fused image‐ new text feature representation $imagenewtextfeat$. The $imagetextfeat$ is passed through the classification layer $f_{clf}$ to obtain the predicted output $output$. The $imagenewtextfeat$ is passed through the same classification layer $f_{clf}$ to obtain the predicted new output $newoutput$.

Five loss functions are calculated during the training stage: $loss_1$ is the classification loss between the true label $y$ and the predicted new output $newoutput$. $loss_2$ is the classification loss between the true label $y$ and the predicted output $output$. Further, to optimize the meta‐ learning module, we calculated the distance between the input and output fea‐ ture embeddings. Hence, $loss_3$ is the distance between the original text feature $textfeat$ and the generated new text feature $newtextfeat$. $loss_4$ is the distance between the fused image‐text feature $imagetextfeat$ and the fused image‐new text feature $imagenewtextfeat$. $loss_5$ is the distance between the predicted output $output$ and the predicted new output $newoutput$.

The total loss is the sum of these five losses as in the Equation 7.

$$TotalLoss = loss_1 + loss_2 + loss_3 + loss_4 + loss_5 \qquad (7)$$

During the testing/ deployment stage, the algorithm takes a test im‐ age $x_v^t$ as input. The test image $x_v^t$ is passed through the ImageEncoder to obtain its encoded representation, which is stored in $x_v$. The encoded test image representation $x_v$ is projected using $Projection_v$ to obtain the image feature representation $imgfeat$. The $imgfeat$ is passed through the TextFeatGen module to generate the corresponding text feature repre‐ sentation $newtextfeat$. The $imgfeat$ and $newtextfeat$ are fused using the ImgTextFuse module to obtain the fused image‐text feature representa‐ tion $imagetextfeat$. The $imagetextfeat$ is passed through the classification layer $f_{clf}$ to obtain the final classification output, which is stored in $output$.

---

**Algorithm 2** Image‐Text Fusion Algorithm

---

**Require:** $x_v, x_t$    Inputs: Images and Text for Training
**Ensure:** $y$    Output: Classification Result

   **Training Stage:**
1: *TextFeatGen: Text Feature Generation Module (See Section ① of ⓒ in Fig. 3)*
2: *ImgTextFuse: Image and Text Feature Fusion Module (See Section ② of ⓒ in Fig. 3)*
3: *$f_{clf}$: Classification Layer*
4: **for** each $(x_v, x_t)$ in *Images, Text* **do**
5:    $x_v \leftarrow$ ImageEncoder$(x_v)$
6:    $x_t \leftarrow$ TextEncoder$(x_t)$
7:    *$imgfeat \leftarrow$* Projection$_v(x_v)$
8:    *$textfeat \leftarrow$* Projection$_t(x_t)$
9:    *$newtextfeat \leftarrow$* TextFeatGen$(imgfeat)$
10:    *$imagetextfeat \leftarrow$* ImgTextFuse$(imgfeat, textfeat)$
11:    *$imagenewtextfeat \leftarrow$* ImgTextFuse$(imgfeat, newtextfeat)$
12:    *$output \leftarrow f_{clf}(imagetextfeat)$*
13:    *$newoutput \leftarrow f_{clf}(imagenewtextfeat)$*
14:    $loss_1 \leftarrow$ ClassificationLoss$(y, newoutput)$
15:    $loss_2 \leftarrow$ ClassificationLoss$(y, output)$
16:    $loss_3 \leftarrow$ distance$(textfeat, newtextfeat)$
17:    $loss_4 \leftarrow$ distance$(imagetextfeat, imagenewtextfeat)$
18:    $loss_5 \leftarrow$ distance$(output, newoutput)$
19:    $loss \leftarrow loss_1 + loss_2 + loss_3 + loss_4 + loss_5$
20: **end for**
   **Testing Stage:**
**Require:** $x_v^t$    Input: Testing Images
21: **for** each $x_v^t$ in *Images* **do**
22:    $x_v \leftarrow$ ImageEncoder$(x_v^t)$
23:    *$imgfeat \leftarrow$* Projection$_v(x_v)$
24:    *$newtextfeat \leftarrow$* TextFeatGen$(imgfeat)$
25:    *$imagetextfeat \leftarrow$* ImgTextFuse$(imgfeat, newtextfeat)$
26:    *$output \leftarrow f_{clf}(imagetextfeat)$*
27: **end for**

---

# 4 | EXPERIMENTS

**Dataset:** As tabulated in Table 1, we adopt the CelebA dataset, for train‐ ing the gender classifier, as each image has 40 true/false attributes – pointy nose, an oval face, gray hair, wearing hat, etc, which can be used to create captions. Additionally, our study entails a comprehensive ex‐ perimental evaluation of widely recognized benchmark datasets such as FairFace, UTKFace, and DiveFace wherein we systematically compare the efficacy of our proposed methodology with existing state‐of‐the‐ art approaches. The images in these datasets vary across age, gender, pose, lighting conditions, and expression. These datasets are discussed in Table 2.

**TABLE 1** Overview of the classification task studied and the datasets used.

| Task | Training Dataset | Test Dataset |
|------|-----------------|--------------|
| Gender Classification | CelebA Liu et al. (2015) | FairFace Kärkkäinen and Joo (2019), UTKFace Zhang et al. (2017a), DiveFace Morales et al. (2021) |

In our experimental framework aimed at generating textual descrip‐ tions for each image, we adopted a dual‐captioning strategy. Specif‐ ically, two captions were generated for each image: one utilizing an image‐to‐text model, namely BLIP (Bootstrapping Language Image Pre‐ training) Li et al. (2022),§(referred to as *Text Generated using BLIP Image to Text model (T1)* in Tables 3, 4, 5, 6). BLIP is a multimodal model in‐ corporating an encoder‐ decoder architecture pre‐trained on a dataset bootstrapped from large‐ scale noisy image‐text pairs. Language Mod‐ eling in BLIP activates the image‐grounded text decoder, optimizing a cross‐entropy loss to generate textual descriptions given an image in an autoregressive manner.

Additionally, the second caption utilizes annotated attributes, denoted as *Text Generated using Attributes* (T2) in Tables 3, 4, 5, and 6. The CelebA dataset encompasses a diverse set of at‐ tributes, including *Arched_Eyebrows, Attractive, Bags_Under_Eyes, Bald, Bangs, Big_Lips, Big_Nose, Black_Hair, Blond_Hair, Blurry, Brown_Hair, Bushy_Eyebrows, Chubby, Double_Chin, Eyeglasses, Goatee, Gray_Hair, Heavy_Makeup, High_Cheekbones, Male, Mouth_Slightly_Open, Mustache, Narrow_Eyes, No_Beard, Oval_Face, Pale_Skin, Pointy_Nose, Reced‐ ing_Hairline, Rosy_Cheeks, Sideburns, Smiling, Straight_Hair, Wavy_Hair, Wearing_Earrings, Wearing_Hat, Wearing_Lipstick, Wearing_Necklace, Wearing_Necktie, Young*. The caption is meticulously crafted to encom‐ pass all annotated attributes associated with each image. Illustrative image captions are presented in Figures 2 & 3.

Employing a dual‐captioning strategy, we aim to investigate the influence of two distinct image captions on visual classification. This approach not only facilitates a comprehensive analysis of the proposed method's versatility and performance but also establishes a robust foundation for evaluating its effectiveness. Figures 2 & 3 showcase a representative example of the generated textual descriptions, offering insights into the quality and diversity of captions produced by our approach.

**Implementation:** Our models are implemented in PyTorch‐Lightning and trained on two NVIDIA RTX $6000$ GPUs. Our visual encoder is built upon ViT‐B/32 Dosovitskiy et al. (2021) architecture, using an input image size of 224×224, a patch size of 32×32, and an embedding dimension of $d = 1024$. Concurrently, BERT$_{base}$ Devlin et al. (2019)is used as our chosen text encoder. During the training phase, empirically, the model is trained for $30$ epochs with a batch size of $128$ using an early stop‐ ping mechanism. We employ the mini‐batch RMSprop optimizer with a weight decay of $5 \times e^{-}4$ for optimization. The learning rate is initial‐ ized at $1 \times e^{-}5$ and undergoes a warm‐up phase to $1 \times e^{-}4$ after every $10$ training epochs, followed by a subsequent decrease using the cosine decay strategy down to $1 \times e^{-}5$. Data augmentation techniques such as random color jittering, random erasing of pixels, and the application of RandAugment Cubuk et al. (2020) are used to enhance robustness. These meticulously chosen architectural and training configurations collectively contribute to the model's ability to learn and represent im‐ age information with text guidance effectively. In our implementation, we used HuggingFace¶ and the Timm repository# Wightman (2019) to obtain the pre‐trained weights of BERT and Vision transformer.

The combined loss function consisting of cross‐entropy, and focal loss Lin et al. (2017) were used as classification loss in the equations 3 and 7. The focal loss aims to reduce and address class imbalance dur‐ ing training, particularly beneficial in tasks like object detection. Focal Loss is a modification of the standard Cross‐Entropy Loss, designed to emphasize the learning of misclassified examples during training. The Cross‐Entropy Loss is defined as:

$$CE(p, y) = \begin{cases} -\log(p) & \text{if } y = 1 \\ -\log(1-p) & \text{otherwise.} \end{cases} \tag{8}$$

Where $y$ is the ground‐truth label, and $p$ is the model's estimated probability for the positive class.

The Focal Loss introduces a modulating factor $(1 - p_t)^\gamma$, which down‐ weights the loss contribution from easy, properly classified examples, and focuses training on the hard negatives:

$$FL(p_t) = -(1 - p_t)^\gamma \log(p_t) \tag{9}$$

Here, $p_t$ is $p$ if $y = 1$, else $1 - p$. The focusing parameter $\gamma \geq 0$ adjusts the weight given to hard examples.

This loss reshaping prevents the overwhelming majority of easy neg‐ atives from dominating the gradient during training, enabling the model to better learn from hard, misclassified cases in scenarios with class imbalance like object detection.

Further, the Information Noise‐Contrastive Estimation (InfoNCE) loss van den Oord et al. (2018), is used as a distance function that aims to optimize the meta‐learning module. We aimed to estimate the mu‐ tual information between a pair of variables by discriminating between each positive pair and its associated $K$ negative pairs. A relevance func‐ tion $f(., .)$ (in this work, the feature embedding) is used to measure the non‐normalized mutual information score between them. For each pos‐ itive sample $(x_+, c)$, it is associated with $K$ random negative samples, denoted as $\{(x^-_1, c), (x^-_2, c), ..., (x^-_K, c)\}$. Then, the InfoNCE loss function $L_K$ is formulated as follows:

$$L_K = -\log \left( \frac{e^{f(x^+, c)}}{e^{f(x^+, c)} + \sum_{j=1}^{K} e^{f(x^-_j, c)}} \right). \tag{10}$$

§ Accessible at https://huggingface.co/Salesforce/blip‐image‐captioning‐base

¶ https://huggingface.co/

# https://github.com/rwightman/pytorch‐image‐models

**T A B L E 2** Details on the datasets used.

| Dataset | Description |
|---|---|
| CelebA Liu et al. (2015) | The CelebFaces Attributes Dataset (CelebA) is a large-scale face attributes dataset with more than 200K celebrity images, each with 40 attribute annotations. The images in this dataset cover large pose variations and background clutter. CelebA has large diversity, large quantity, and rich annotations. CelebA also includes a recommended subject-disjoint split into train, validation, and test. |
| FairFace Kärkkäinen and Joo (2019) | The FairFace dataset consists of 108,501 images, with an emphasis on balanced race composition. The dataset is labeled with seven race groups: White, Black, Indian, East Asian, Southeast Asian, Middle Eastern, and Latino Hispanic, across male and female genders and age groups ranging from 0–9, 10–19, 20–29, 30–39, 40–49, and 50+. |
| UTKFace Zhang et al. (2017a) | The UTKFace dataset is a facial image dataset spanning ages from 0 to 116 years. It contains over 20,000 face images annotated with age, gender, and ethnicity (White, Black, Asian, Indian, and Others, which includes Hispanic, Latino, and Middle Eastern). Due to the vagueness of the "Other" category, we excluded it from this study. The dataset includes significant variations in pose, expression, illumination, occlusion, and resolution. |
| DiveFace Morales et al. (2021) | The DiveFace dataset contains a total of 139,677 images with gender and race annotations equally distributed among three ethnic groups: East Asian, Sub-Saharan and South Indian, and Caucasian. |

In Equation 7, the InfoNCE loss ($L_K$) is denoted as the distance func-tion, contributing to the overall optimization objective.

**Evaluation Metrics:** To conduct a comprehensive assessment of all mod-els' performance and quantify bias following prior research Lin et al. (2022), Singh et al. (2022b), we employed standard evaluation metrics Krishnan and Rattani (2023) used for the bias evaluation of the facial attribute classifiers as follows:

*Overall classification accuracy, Degree of Bias (DoB) represented by the standard deviation of accuracy across specified demographics, and the ratio of maximum and minimum accuracy values explained as follows:*

1. DoB (standard deviation of accuracy across demographics) assesses variability of model performance among sub-groups. Low DoB indi-cates consistent classification accuracy across demographics, hence, indicating reduced bias.
2. Ratio of maximum and minimum accuracy values indicates dispar-ities across demographic subgroups. The ratio of unity signifies consistent performance across diverse demographic sub-groups, indicating fairness and unbiased treatment.

In this study, particular emphasis was placed on DoB and the max-min accuracy ratio, as they are crucial measures for assessing fairness and demographic parity.

**Results:** Our proposed text-guided facial-image-based gender classifi-cation approach obtains consistently superior performance compared to established baselines across all evaluated datasets. This claim is sub-stantiated by the results presented in Tables 3, 4, 5, and 6, which detail the classification accuracy and overall bias obtained on intra- and cross-dataset evaluations of our method against several benchmark methods. These baselines include:

- Image-only operation (See Section (A) in the Figure 2): A standard gender classifier trained solely on facial images.

- CLIP-based approach: A method utilizing CLIP Radford et al. (2021) with both CelebA dataset images and corresponding text descrip-tions (generated by BLIP or extracted from image attributes) during training, employing the InfoNCE loss function for representation learning.
- TandemNet Zhang et al. (2021): A method leveraging a dual-attention mechanism for enhanced visual and semantic information alignment.

Tables 3, 4, 5, and 6 provide a comprehensive comparison across various datasets, demonstrating the clear advantage of our text-guided approach in both accuracy and bias reduction.

Our intra-dataset evaluation on CelebA (refer Table 3) meticulously compares gender classification performance across baseline and pro-posed methods. Baselines, including image-only and recent text-guided vision approaches Zhang et al. (2021), Radford et al. (2021), achieve commendable accuracy ($97.33\%$-$99.15\%$ for males and $97.87\%$-$99.0\%$ for females). Notably, Max/Min ratios ($1.007$-$1.019$) and Degree of Bias (DoB) ($0.515$-$1.291$) vary.

Our proposed methods, ITM-guided and Image-Text fusion strate-gies, employ two types of text inputs: BLIP-generated and attribute-based. These methods demonstrate competitive results. Notably, the ITM-guided strategy with attribute-based text achieves the lowest Maximum/Minimum ratio ($1.006$) and Degree of Bias (DoB) ($0.441$), outperforming the baseline methods. Additionally, it obtains an over-all accuracy of 98.567%, higher than the baselines. Furthermore, the Image-Text fusion strategy with attribute-based text obtains the highest overall accuracy of 98.598%, highlighting its effectiveness.

From Table 3, it is evident that the proposed ITM-guided method with attribute-based text outperforms the baselines, particularly by reducing the Difference of Bias by 15% and modestly improving the overall ac-curacy compared to the best baseline. This reinforces the efficacy of text-guided classification approaches in mitigating bias and improving accuracy.

**TABLE 3** Gender Classification Accuracy (%) on the CelebA test set across different demographics. M: Male, F: Female. Max/Min indicates the ratio of maximum to minimum classification accuracy across gen- der and ethnicity; Overall represents the average classification accuracy (%); DoB refers to the standard deviation across genders. The top per- formance results are highlighted in bold. T1: Text generated using BLIP Image-to-Text Model; T2: Text generated using Attributes. References: Zhang et al. (2021): TandemNet; Radford et al. (2021): CLIP.

| Method | M | F | Max/Min ↓ | Overall ↑ | DoB ↓ |
|---|---|---|---|---|---|
| **Baseline** | | | | | |
| Image Only | 97.33 | 99.151 | 1.019 | 98.447 | 1.288 |
| TandemNet - T1 Zhang et al. (2021) | 97.99 | 98.718 | 1.007 | 98.437 | 0.515 |
| TandemNet - T2 Zhang et al. (2021) | 97.874 | 98.734 | 1.009 | 98.402 | 0.608 |
| CLIP - T1 Radford et al. (2021) | 97.174 | 99.000 | 1.019 | 98.294 | 1.291 |
| CLIP - T2 Radford et al. (2021) | 98.0946 | 98.873 | 1.008 | 98.572 | 0.550 |
| **Proposed Method** | | | | | |
| ITM Guided - T1 | 97.369 | 99.159 | 1.018 | 98.467 | 1.266 |
| ITM Guided - T2 | 98.185 | 98.808 | **1.006** | 98.567 | **0.441** |
| Image-Text Fusion - T1 | 97.446 | 99.281 | 1.019 | 98.572 | 1.298 |
| Image-Text Fusion - T2 | 97.343 | 99.388 | 1.021 | **98.598** | 1.446 |

The cross-dataset evaluation, as presented in Tables 4,5, & 6), demonstrates the superiority of the proposed method across various datasets. We evaluated the overall accuracy, the degree of bias, and the ra- tio of maximum to minimum accuracies across different gender-racial groups. The results highlight the effectiveness of the proposed method in achieving higher overall accuracy, mitigating bias, and maintaining consistent performance across diverse demographic groups when com- pared to other methods evaluated on these datasets.

On DiveFace (refer Table 4), baselines showed accuracy between 95.5% (Sub-Saharan & South Indian) and 99.4% (White) for males and 76.9% (Sub-Saharan & South Indian) and 99.2% (East Asian) for females, with Max/Min ratios ranging from $1.038$ to $1.292$ and DoB ranging from $1.444$ to $7.964$.

Applying the proposed Image-Text Fusion with BLIP-generated text achieves the lowest Max/Min ratio (1.033), highest overall accuracy (97.218%), and lowest DoB (1.326), demonstrating effectiveness in mit- igating bias and improving accuracy across diverse gender-racial groups. While the ITM-guided method with BLIP-generated text outperformed most baselines, it did not surpass the TandemNet with BLIP-generated text in terms of overall accuracy.

The Image-Text Fusion with BLIP-generated text exhibited a mod- est improvement in overall accuracy compared to the best baseline (97.218% vs. 97.14%). However, its true superiority lies in achieving a 9% reduction in DoB compared to the best baseline, confirming the ef- fectiveness of text-guided classification in mitigating bias and improving accuracy on the DiveFace dataset.

The UTKFace dataset (Refer Table 5) exhibits similar trends to Dive- Face. Baselines, encompassing both image-only and recent text-guided approaches Zhang et al. (2021), Radford et al. (2021), show varied accuracy across races: 65%-98.27% for males and 75%-97% for fe- males. Notably, Max/Min ratios range from 1.353 to 1.506, and DoB metrics range from 8.2 to 9.257. Applying, Image-Text Fusion with

BLIP-generated text reduced the bias up to 49%, and improved overall classification up to 7% when compared over the baselines.

The FairFace dataset (Refer Table 6) reveals similar trends to Dive- Face and UTKFace. Baseline methods, including both image-only and recent text-guided approaches Zhang et al. (2021), Radford et al. (2021), demonstrate commendable accuracy levels ranging from 78% (East Asian) to 93.891% (Indian) for males and 54.74% (Black) to 92.5% (East Asian) for females. However, they exhibit variations in Max/Min ra- tios (1.185 to 1.706) and DoB metrics (4.82 to 12.43). Our proposed methods (ITM- guided and Image-Text Fusion with two text inputs) showcase competitive performance. Notably, Image-Text Fusion with BLIP-generated text outperformed the reducing the degree of bias by up to 13%, and an improvement of overall accuracy by up to 7%.

The cross-dataset evaluation confirmed the superiority of the proposed Image-Text Fusion technique, especially when using BLIP- generated text. This approach demonstrated significant improvements in mitigating bias, with up to 49% reduction in the degree of bias, and up to 7% improvement in overall accuracy compared to other methods.

The superior performance of the Image-Text Fusion approach can be attributed to its end-to-end optimization for the classification task. This allows the model to learn the optimal way to combine image and text modalities, leading to better performance compared to using image- text matching guidance, which may be an auxiliary task not directly optimized for classification.

Furthermore, the better performance achieved with BLIP-generated text, compared to text generated from annotated attributes, is likely due to the rich contextual information present in the natural language descriptions from BLIP. Annotated attributes can often be sparse and in- complete, limiting their effectiveness in guiding the classification task. In contrast, BLIP's pretraining on large-scale image-text data enables it to generate more informative and relevant text descriptions, contributing to improved performance.

Overall, these results reinforce the effectiveness of text-guided classification approaches in improving gender classification accuracy and mitigating bias across diverse datasets.

**Comparative Analysis with SOTA:** We conducted a comparative anal- ysis between our proposed bias mitigation technique and state-of- the-art (SOTA) methods based on CLIP Radford et al. (2021), multi- tasking Das et al. (2018), and self-consistency Krishnan and Rattani (2023) for gender classification. The evaluation was performed on three distinct datasets: Fairface, UTKFace, and DiveFace. The algorithms un- derwent training using the CelebA dataset. Our selected model for this comparative study is the Image-Text Fusion approach.

For the comparative analysis, we used overall classification accuracy, the ratio of maximum and minimum accuracy values, and the DoB as metrics shown in Table 7. Our proposed Image-Text Fusion approach achieved the best performance across multiple evaluation metrics on Fairface, UTKFace, and DiveFace datasets.

On FairFace, our approach reduced DoB by 12.4% compared to CLIP. On UTKFace, it achieved 0.54% higher overall accuracy and **T A B L E 4** Gender Classification Accuracy (%) on DiveFace test set across different demographics. M: Male, F: Female. Max/Min represents the ratio of maximum to minimum classification accuracy across gender and ethnicity; Overall indicates the average classification accuracy (%); DoB is the standard deviation across all values. The top performance results are highlighted in bold. T1: Text generated using BLIP Image-to-Text Model; T2: Text generated using Attributes. References: Zhang et al. (2021): TandemNet; Radford et al. (2021): CLIP.

| Method | East Asian | | Sub-Saharan & South Indian | | White | | Max/Min ↓ | Overall ↑ | DoB ↓ |
|---|---|---|---|---|---|---|---|---|---|
| | M | F | M | F | M | F | | | |
| **Baseline** | | | | | | | | | |
| Image Only | 96.374 | 96.231 | 97.675 | 86.670 | 99.116 | 92.426 | 1.143 | 94.753 | 4.147 |
| TandemNet - T1 Zhang et al. (2021) | 95.648 | 99.167 | 95.488 | 96.532 | 98.895 | 97.050 | 1.039 | 97.149 | 1.444 |
| TandemNet - T2 Zhang et al. (2021) | 96.600 | 95.924 | 97.174 | 88.243 | 99.426 | 92.162 | 1.127 | 94.924 | 3.683 |
| CLIP - T1 Radford et al. (2021) | 97.371 | 95.136 | 98.359 | 82.793 | 99.426 | 90.136 | 1.200 | 93.867 | 5.795 |
| CLIP - T2 Radford et al. (2021) | 97.000 | 92.682 | 98.500 | 76.892 | 99.337 | 86.438 | 1.292 | 91.798 | 7.964 |
| **Proposed Method** | | | | | | | | | |
| ITM Guided - T1 | 95.330 | 99.036 | 95.168 | 96.620 | 99.116 | 96.610 | 1.042 | 97.000 | 1.584 |
| ITM Guided - T2 | 97.189 | 97.195 | 97.310 | 89.865 | 99.580 | 93.000 | 1.108 | 95.694 | 3.248 |
| Image-Text Fusion - T1 | 95.830 | 98.948 | 96.000 | 96.306 | 99.028 | 97.094 | **1.033** | **97.218** | **1.326** |
| Image-Text Fusion - T2 | 95.014 | 98.773 | 96.080 | 94.680 | 99.116 | 95.773 | 1.047 | 96.591 | 1.741 |

**T A B L E 5** Gender Classification Accuracy (%) on UTKFace test set across different demographics. M: Male, F: Female. Max/Min indicates the ratio of maximum to minimum classification accuracy among gender and ethnicity; Overall represents the average classification accuracy (%); DoB refers to the standard deviation across all values. The top performance results are highlighted in bold. T1: Text generated using BLIP Image-to-Text Model; T2: Text generated using Attributes. References: Zhang et al. (2021): TandemNet; Radford et al. (2021): CLIP.

| Method | Asian | | Black | | Indian | | White | | Max/Min ↓ | Overall ↑ | DoB ↓ |
|---|---|---|---|---|---|---|---|---|---|---|---|
| | M | F | M | F | M | F | M | F | | | |
| **Baseline** | | | | | | | | | | | |
| Image Only | 68.79 | 93.513 | 98.268 | 83.640 | 91.150 | 88.304 | 90.494 | 84.130 | 1.428 | 87.893 | 8.300 |
| TandemNet - T1 Zhang et al. (2021) | 68.153 | 91.892 | 98.268 | 80.910 | 92.478 | 84.210 | 90.859 | 82.391 | 1.442 | 86.982 | 8.741 |
| TandemNet - T2 Zhang et al. (2021) | 72.611 | 87.567 | 98.268 | 75.000 | 90.265 | 81.871 | 90.310 | 78.700 | 1.353 | 85.026 | 8.200 |
| CLIP - T1 Radford et al. (2021) | 67.516 | 96.757 | 96.970 | 92.273 | 92.920 | 94.152 | 92.139 | 92.174 | 1.436 | 91.534 | 8.923 |
| CLIP - T2 Radford et al. (2021) | 64.968 | 94.594 | 97.835 | 85.450 | 90.625 | 90.058 | 89.031 | 87.609 | 1.506 | 88.293 | 9.257 |
| **Proposed Method** | | | | | | | | | | | |
| ITM Guided - T1 | 70.060 | 94.054 | 96.537 | 88.182 | 92.478 | 92.982 | 92.322 | 90.435 | 1.378 | 90.578 | 7.740 |
| ITM Guided - T2 | 66.240 | 93.513 | 97.402 | 87.273 | 90.265 | 90.643 | 89.945 | 87.174 | 1.470 | 88.575 | 8.725 |
| Image-Text Fusion - T1 | 84.076 | 89.189 | 96.970 | 87.727 | 96.018 | 88.890 | 95.247 | 91.087 | **1.153** | **92.080** | **4.258** |
| Image-Text Fusion - T2 | 73.885 | 93.513 | 97.403 | 85.450 | 92.478 | 87.134 | 92.505 | 86.304 | 1.318 | 89.349 | 6.760 |

**T A B L E 6** Gender Classification Accuracy (%) on FairFace test set across different demographics. M: Male, F: Female. Max/Min is the ratio of maximum to minimum classification accuracy among gender and ethnicity; Overall represents the average classification accuracy (%); DoB is the standard deviation across all values. The top performance results are highlighted in bold. T1: Text generated using BLIP Image-to-Text Model; T2: Text generated using Attributes. References: Zhang et al. (2021): TandemNet; Radford et al. (2021): CLIP.

| Method | Black | | East Asian | | Indian | | Latino Hispanic | | Middle Eastern | | Southeast Asian | | White | | Max/Min ↓ | Overall ↑ | DoB ↓ |
|---|---|---|---|---|---|---|---|---|---|---|---|---|---|---|---|---|---|
| | M | F | M | F | M | F | M | F | M | F | M | F | M | F | | | |
| **Baseline** | | | | | | | | | | | | | | | | | |
| Image Only | 89.467 | 64.220 | 79.794 | 87.970 | 88.524 | 83.735 | 88.524 | 83.735 | 92.743 | 78.283 | 81.088 | 83.823 | 89.750 | 82.700 | 1.444 | 84.324 | 7.833 |
| TandemNet - T1 Zhang et al. (2021) | 90.087 | 58.610 | 84.170 | 81.760 | 93.891 | 60.419 | 91.047 | 76.144 | 93.727 | 70.454 | 85.306 | 80.147 | 91.440 | 77.674 | 1.602 | 81.782 | 11.090 |
| TandemNet - T2 Zhang et al. (2021) | 90.706 | 54.740 | 83.140 | 80.466 | 93.360 | 57.405 | 90.668 | 73.373 | 92.866 | 67.172 | 85.034 | 75.588 | 91.711 | 72.900 | 1.706 | 80.036 | 12.430 |
| CLIP - T1 Radford et al. (2021) | 80.421 | 80.107 | 81.596 | 92.497 | 88.446 | 84.404 | 87.390 | 90.602 | 92.000 | 88.384 | 79.184 | 93.823 | 90.285 | 89.720 | 1.185 | 87.164 | 4.817 |
| CLIP - T2 Radford et al. (2021) | 86.121 | 66.889 | 78.121 | 88.228 | 90.040 | 74.574 | 86.507 | 85.181 | 91.636 | 83.840 | 80.680 | 88.235 | 89.572 | 83.074 | 1.370 | 83.974 | 6.553 |
| **Proposed Method** | | | | | | | | | | | | | | | | | |
| ITM Guided - T1 | 81.536 | 77.704 | 82.883 | 91.203 | 89.376 | 79.948 | 87.768 | 89.277 | 92.500 | 87.626 | 80.136 | 91.323 | 91.000 | 88.473 | 1.190 | 86.642 | 4.800 |
| ITM Guided - T2 | 80.793 | 73.300 | 73.359 | 90.686 | 87.384 | 80.472 | 83.480 | 89.036 | 90.529 | 86.616 | 74.014 | 90.441 | 88.948 | 85.462 | 1.237 | 84.019 | 6.265 |
| Image-Text Fusion - T1 | 85.000 | 77.036 | 86.615 | 90.168 | 90.704 | 82.438 | 87.894 | 90.723 | 92.743 | 88.890 | 82.449 | 90.588 | 90.820 | 89.512 | 1.204 | **87.670** | **4.220** |
| Image-Text Fusion - T2 | 84.387 | 72.096 | 77.349 | 90.039 | 89.510 | 74.443 | 86.885 | 87.000 | 91.390 | 84.850 | 78.095 | 90.147 | 89.216 | 84.943 | 1.268 | 84.491 | 6.072 |

52.3% lower DoB than CLIP. On DiveFace, it outperformed Multi- Tasking with 0.93% higher accuracy and 30.3% lower DoB. Our method consistently demonstrated modestly improved performance across all datasets, achieving the lowest bias (DoB), most consistent performance (lowest Maximum/Minimum Ratio), and highest overall accuracy com- pared to the state-of-the-art methods.

Across all three datasets, our Image-Text Fusion approach consistently demonstrated the lowest Max/Min ratio, indicating robust performance across subgroups. It also achieved the highest overall accuracy, highlighting its efficacy in improving gender classification performance. Most importantly, it attained lower DoB scores compared to the state-of-the-art methods, affirming its strength in mitigating biases related to gender classification. This could be because, In image-text fusion, the image and text features are combined/fused, allowing the model to leverage the complementary information from both modal- ities jointly during classification. Hence, quantitative results on three distinct datasets validate that our proposed approach exhibits robust- ness, high accuracy, and reduced bias for gender classification tasks.

**Ablation Study:** Our experimental results underscore the crucial impact of text caption on both generalization and fairness of the image-based vision classifier. To explore this phenomenon, we conducted experi- ments to identify the optimal attributes from the set of 40 attributes from the CelebA facial image dataset. The Image-Text fusion technique was employed for this study because of its superior performance over ITM.

CelebA attribute list include: *Arched_Eyebrows, Attractive, Bags_Under_Eyes, Bald, Bangs, Big_Lips, Big_Nose,Black_Hair, Blond_Hair, Blurry, Brown_Hair, Bushy_Eyebrows, Chubby, Double_Chin, Eye - glasses, Goatee, Gray_Hair, Heavy_Makeup, High Cheekbones, Male, Mouth_Slightly_Open, Mustache, Narrow_Eyes, No_Beard, Oval_Face, Pale_Skin, Pointy_Nose, Receding_Hairline, Rosy_Cheeks, Sideburns, Smiling, Straight_Hair, Wavy_Hair, Wearing_Earrings, Wearing_Hat, Wearing_Lipstick, Wearing_Necklace, Wearing_Necktie, Young*.

**Human-Guided Attribute Selection:** In this experimental investigation, we conducted a human-centered study wherein a participant was tasked with selecting three unique sets of attributes that they deemed as an optimal descriptor of the facial images from a predefined pool of 40 attributes. The textual captions were then generated based on each set of attributes. Table 8 presents different sets of human-selected at- tributes used to generate text descriptions for images. These attribute combinations were curated to enable a comparative evaluation of their effectiveness in mitigating bias and enhancing the performance of gen- der classification models. By analyzing the results obtained with text descriptions derived from these diverse attribute sets, the study aims to assess the impact and utility of attribute-based text representations in addressing bias and improving the accuracy of gender classification tasks across diverse demographic groups. The results of this experiment, including the evaluation metrics, are presented in the following Tables 9, 10 & 11, tabulates the outcome of this experiment on the FairFace, UTKFace, and DiveFace datasets.

This study aims to determine the optimal number of attributes required for generating text descriptions that can outperform the BLIP- generated text in mitigating bias and improving gender classification accuracy. The goal is to investigate the relevance and effectiveness of using attribute-based text representations in addressing bias and en- hancing the performance of gender classification models. By analyzing the performance achieved with different numbers of attributes, this re-search sheds light on the role and significance of attribute information in mitigating biases and improving fairness in gender classification tasks across gender-racial groups.

Table 9 highlights the impact of different attribute sets (T1 to T5) on image caption generation, particularly focusing on races and gen- ders. Observations reveal variations in performance across attribute sets, with Text generated using BLIP (T1) consistently demonstrating su- perior performance. Text generated using BLIP outperforms others in mitigating biases associated with different attributes, emphasizing its effectiveness in generating captions with reduced bias.

Similar trends are observed in Table 10, the text generated using BLIP (T1) consistently outperforms other attribute sets, reaffirming its efficacy in generating captions with minimized biases across diverse at- tributes. Table 11 analyzes races and genders using a different set of attributes. Once again, Text generated using BLIP (T1) stands out by achieving the highest overall performance and effectively minimizing biases across different attributes.

Based on these observations, while the BLIP-generated text (T1) con- sistently outperformed other attribute-based text generation methods in mitigating bias and achieving higher overall performance, the more de- scriptive attribute-based texts (T3, T4, and T5) did not perform as well. This raises an interesting point: despite including more detailed attribute information, these texts were not as effective as the BLIP-generated text. This could be because of multiple potential reasons. Firstly, even though the additional attributes provide more descriptive information, they may still be incomplete or sparse representations of the visual con- tent, limiting their effectiveness in capturing the nuances required for accurate and unbiased gender classification. Secondly, some of the addi- tional attributes included in T3, T4, and T5 may not be directly relevant or influential for the gender classification task, potentially introducing noise or irrelevant information that could hinder performance. And fi- nally, the specific combinations of attributes used in T3, T4, and T5 may not have been optimal or complementary for mitigating bias and improving classification accuracy.

These observations highlight the limitations of relying solely on attribute-based text representations and suggest the potential advan- tages of using more sophisticated vision-language models like BLIP for generating unbiased and accurate text descriptions.

These findings motivate further research into more effective ways of incorporating attribute information or combining it with powerful vision-language models to improve performance and fairness. Addi- tionally, the comprehensive evaluation and comparison of different text generation approaches provide a holistic understanding of their strengths and weaknesses, contributing valuable insights for mitigating bias in gender classification tasks using text-based guidance.

**Prompt-Guided Attribute Selection:** In this experiment, we employed a prompt-guided approach for attribute selection, utilizing ChatGPT to choose three distinct sets of attributes from a pool of 40. The prompts used for attribute selection are detailed in Table 12, named P1, P2 & P3.

**T A B L E 7** Comparative Analysis. A: Multi‑Tasking Das et al. (2018), B: CLIP Radford et al. (2021), C: Self‑Consistency Krishnan and Rattani (2023). Ours: Image‑Text Fusion Approach. Top performance results are highlighted in bold.

| Method | FairFace | | | UTKFace | | | DiveFace | | |
|---|---|---|---|---|---|---|---|---|---|
| | Max/Min ↓ | Overall ↑ | DoB ↓ | Max/Min ↓ | Overall ↑ | DoB ↓ | Max/Min ↓ | Overall ↑ | DoB ↓ |
| A | 1.230 | 84.54 | 5.92 | 1.507 | 88.47 | 9.53 | 1.062 | 96.29 | 1.90 |
| B | **1.185** | 87.16 | 4.82 | 1.436 | 91.53 | 8.92 | 1.200 | 93.87 | 5.80 |
| C | 1.450 | 83.13 | 8.04 | 1.415 | 87.88 | 8.13 | 1.130 | 95.53 | 3.95 |
| **Ours** | 1.204 | **87.67** | **4.22** | **1.153** | **92.08** | **4.26** | **1.033** | **97.22** | **1.33** |

**T A B L E 8** Text Generation via Selected Attributes.

| Caption | How is it Generated? |
|---|---|
| T1 | Text generated using BLIP Image‑to‑Text Model. |
| T2 | Text generated using the complete set of annotated attributes. |
| T3 | Text generated based on specific attributes: Young, Attractive, Male, Smiling, Black Hair, Blonde Hair, Brown Hair, Eyeglasses, Pale Skin, and Blurry. |
| T4 | Text generated based on attributes: Young, Attractive, Smiling, Male, and Beard. |
| T5 | Text generated based on attributes: Attractive, Smiling, Male, High Cheekbones, and Lipstick. |

Subsequently, image captions were generated based on each set of selected attributes. The evaluation results are presented in the following tables 13, 14 & 15, illustrating the model's performance on the FairFace, UTKFace, and DiveFace datasets. This experiment aims to assess the impact of prompt‑guided attribute selection on the generated captions and evaluate the model's adaptability to alleviate bias and improve clas‑ sification accuracy. Further, prompts used to obtain the attributes for generating image captions are shown in Table 12.

Tables 13, 14, and 15 present a thorough examination of the performance of different prompt‑generated attribute sets in generating image captions. The experiments were designed to identify attribute sets capable of producing captions that effectively alleviate bias across gender‑racial groups.

Table 13 provides insights into the impact of different attribute sets on image caption generation, particularly focusing on various races and genders. The observations reveal performance variations across attribute sets, with Text generated using BLIP (T1) consistently demon‑ strating superior performance. This attribute set outperforms others in mitigating biases associated with different attributes, emphasizing its effectiveness in generating captions with reduced bias.

Similar trends are observed in Table 14, which also emphasizes races and genders. Text generated using BLIP (T1) consistently outperforms other attribute sets, reaffirming its efficacy in generating captions with minimized biases across diverse attributes.

Table 15 presents an analysis of races and genders using a different set of attributes. Once again, Text generated using BLIP (T1) stands out by achieving the highest overall performance and effectively minimizing biases across different attributes.

While the attribute‑based approaches demonstrate their potential in mitigating bias, the consistent outperformance of BLIP suggests that its learned representations and end‑to‑end training may better capture the nuances required for unbiased and accurate gender classification across diverse gender‑racial groups.

The experiments underscore the importance of exploring more sophisticated vision‑language models like BLIP, which could potentially leverage their powerful representation learning capabilities to address bias and improve performance in various computer vision tasks involving visual and textual modalities. However, the dynamic nature of biases and the evolving landscape of image datasets necessitate ongoing and in‑depth studies to refine attribute selection strategies for enhanced accuracy and bias mitigation.

**Future Research:** A critical avenue for future research involves exploring an expanded range of attributes, thereby ensuring a nuanced understanding of attribute set variability and its impact on mitigating biases. Additionally, to enhance the applicability and generalizability of findings, future research should encompass a broader spectrum of datasets, facilitating a universal assessment of attribute set efficacy in bias mitigation.

Further strides can be made through the exploration of advanced algorithms tailored for attribute selection and caption generation, potentially leading to enhanced bias reduction. Novel techniques and models warrant thorough investigation to uncover avenues for continual improvement. Furthermore, the integration of user feedback mechanisms into attribute selection processes stands as a promising area for future exploration, contributing to the refinement of attribute sets based on real‑time user preferences and perceptions. Lastly, addressing the adaptability of these systems to dynamic data distributions remains an important direction for future research. These outlined directions for future research align with the imperative to advance the understanding and effectiveness of attribute sets in mitigating biases within the domain of text‑guided visual models.

**T A B L E 9** Gender Classification Accuracy (%) of Image‑Text Fusion technique on FairFace test set across different demographics. M: Male, F: Female. Max/Min is the ratio of maximum and minimum classification accuracy values among gender and ethnicity; Overall is the average accuracy, and DoB refers to the standard deviation. The top performance results are highlighted in bold. T1: Text Generated using BLIP Image‑to‑Text Model; T2: Text generated using all annotated attributes; T3 to T5: Text generated using selected attributes.

| Race | Black | | East Asian | | Indian | | Latino Hispanic | | Middle Eastern | | Southeast Asian | | White | | Max/Min ↓ | Overall ↑ | DoB ↓ |
|---|---|---|---|---|---|---|---|---|---|---|---|---|---|---|---|---|---|
| Gender | M | F | M | F | M | F | M | F | M | F | M | F | M | F | | | |
| Image‑Text Fusion ‑ T1 | 85.0 | 77.0 | 86.6 | 90.2 | 90.7 | 82.4 | 87.9 | 90.7 | 92.7 | 88.9 | 82.4 | 90.6 | 90.8 | 89.5 | **1.204** | **87.67** | **4.22** |
| Image‑Text Fusion ‑ T2 | 84.4 | 72.1 | 77.3 | 90.0 | 89.5 | 74.4 | 86.9 | 87.0 | 91.4 | 84.8 | 78.1 | 90.1 | 89.2 | 84.9 | 1.268 | 84.491 | 6.072 |
| Image‑Text Fusion ‑ T3 | 79.9 | 77.4 | 75.4 | 91.9 | 87.8 | 80.3 | 85.9 | 90.1 | 90.9 | 88.9 | 76.1 | 91.3 | 87.7 | 89.0 | 1.218 | 85.228 | 5.800 |
| Image‑Text Fusion ‑ T4 | 84.1 | 75.7 | 77.5 | 92.1 | 89.5 | 82.0 | 86.6 | 90.1 | 91.9 | 88.1 | 77.1 | 92.8 | 88.9 | 89.3 | 1.226 | 86.236 | 5.656 |
| Image‑Text Fusion ‑ T5 | 83.3 | 73.0 | 79.2 | 88.5 | 90.6 | 74.3 | 88.5 | 85.4 | 92.7 | 81.8 | 81.6 | 87.4 | 90.1 | 84.9 | 1.270 | 84.701 | 5.720 |

**T A B L E 10** Gender Classification Accuracy (%) of Image‑Text Fusion technique on UTKFace test set across different demographics. M: Male, F: Female. Max/Min is the ratio of maximum to minimum classification accuracy among gender and ethnicity; Overall represents the average accuracy, and DoB refers to the standard deviation. T1: Text Generated using BLIP Image‑to‑Text Model; T2: Text generated using all annotated attributes; T3 to T5: Text generated using selected attributes.

| Method | Asian | | Black | | Indian | | White | | Max/Min ↓ | Overall ↑ | DoB ↓ |
|---|---|---|---|---|---|---|---|---|---|---|---|
| | M | F | M | F | M | F | M | F | | | |
| Image‑Text Fusion ‑ T1 | 84.076 | 89.189 | 96.970 | 87.727 | 96.018 | 88.890 | 95.247 | 91.087 | **1.153** | **92.080** | **4.258** |
| Image‑Text Fusion ‑ T2 | 73.885 | 93.513 | 97.403 | 85.450 | 92.478 | 87.134 | 92.505 | 86.304 | 1.318 | 89.349 | 6.760 |
| Image‑Text Fusion ‑ T3 | 66.242 | 94.594 | 97.403 | 85.000 | 92.035 | 89.474 | 91.590 | 86.960 | 1.470 | 88.895 | 9.000 |
| Image‑Text Fusion ‑ T4 | 75.159 | 93.513 | 97.403 | 85.000 | 93.363 | 91.813 | 92.870 | 89.565 | 1.300 | 90.623 | 6.473 |
| Image‑Text Fusion ‑ T5 | 75.796 | 91.350 | 96.970 | 84.091 | 93.363 | 89.474 | 93.053 | 85.870 | 1.279 | 89.440 | 6.250 |

**T A B L E 11** Gender Classification Accuracy (%) of Image‑Text Fusion technique on DiveFace test set across different demographics. M: Male, F: Female. Max/Min is the ratio of maximum and minimum classification accuracy among gender and ethnicity; Overall represents the average accuracy, and DoB refers to the standard deviation. The top performance results are highlighted in bold. T1: Text Generated using BLIP Image‑to‑Text Model; T2: Text generated using all annotated attributes; T3 to T5: Text generated using selected attributes.

| Method | East Asian | | Sub‑Saharan and South Indian | | White | | Max/Min ↓ | Overall ↑ | DoB ↓ |
|---|---|---|---|---|---|---|---|---|---|
| | M | F | M | F | M | F | | | |
| Image‑Text Fusion ‑ T1 | 95.83 | 98.95 | 96.00 | 96.31 | 99.03 | 97.09 | **1.033** | **97.218** | **1.326** |
| Image‑Text Fusion ‑ T2 | 95.01 | 98.77 | 96.08 | 94.68 | 99.12 | 95.77 | 1.047 | 96.591 | 1.741 |
| Image‑Text Fusion ‑ T3 | 95.56 | 99.08 | 95.35 | 96.31 | 99.20 | 97.62 | 1.040 | 97.209 | 1.562 |
| Image‑Text Fusion ‑ T4 | 96.28 | 98.69 | 97.04 | 93.87 | 99.34 | 95.77 | 1.058 | 96.844 | 1.822 |
| Image‑Text Fusion ‑ T5 | 97.05 | 97.59 | 96.95 | 93.42 | 99.60 | 95.11 | 1.066 | 96.628 | 1.942 |

**T A B L E 12** Prompts and ChatGPT responses.

| | Prompt | ChatGPT Response |
|---|---|---|
| P1 | Select the optimal attributes from the provided list that accurately characterize a facial image. Please choose the definitive attributes from the given list. **List:** CelebA attribute list. | Attractive, High_Cheekbones, Smiling, Young, Blurry. |
| P2 | Select the optimal fixed attributes that primarily emphasize the facial structure. It is crucial that these chosen attributes do not introduce any biases during model training. **List:** CelebA attribute list. | Arched_Eyebrows, High_Cheekbones, Oval_Face, Pointy_Nose, Mouth_Slightly_Open, Narrow_Eyes. |
| P3 | Select the attributes that most effectively depict the features and characteristics of a facial image. Please choose the definitive attributes from the given list. **List:** CelebA attribute list. | Arched_Eyebrows, Blurry, Mouth_Slightly_Open, Oval_Face, Pointy_Nose, Rosy_Cheeks. |

**T A B L E** 13 Gender Classification Accuracy (%) of Image‑Text Fusion technique on FairFace test set across different demographics. M: Male, F: Female. Max/Min represents the ratio of maximum to minimum classification accuracy among gender and ethnicity; Overall is the average accuracy, and DoB refers to the standard deviation. T1: Text Generated using BLIP Image‑to‑Text Model; T2: Text generated using all annotated attributes; P1 to P3 are Text generated using attributes obtained from the prompts in Table 12.

| Method | Black | | East Asian | | Indian | | Latino Hispanic | | Middle Eastern | | Southeast Asian | | White | | Max/Min ↓ | Overall ↑ | DoB ↓ |
|---|---|---|---|---|---|---|---|---|---|---|---|---|---|---|---|---|---|
| | M | F | M | F | M | F | M | F | M | F | M | F | M | F | | | |
| Image‑Text Fusion ‑ T1 | 85.0 | 77.0 | 86.6 | 90.2 | 90.7 | 82.4 | 87.9 | 90.7 | 92.7 | 88.9 | 82.4 | 90.6 | 90.8 | 89.5 | **1.204** | **87.67** | **4.22** |
| Image‑Text Fusion ‑ T2 | 84.4 | 72.1 | 77.3 | 90.0 | 89.5 | 74.4 | 86.9 | 87.0 | 91.4 | 84.8 | 78.1 | 90.1 | 89.2 | 84.9 | 1.268 | 84.491 | 6.072 |
| Image‑Text Fusion ‑ P1 | 82.3 | 66.9 | 79.0 | 89.4 | 89.8 | 74.1 | 88.3 | 84.8 | 91.6 | 81.8 | 80.0 | 89.0 | 90.1 | 82.3 | 1.370 | 83.803 | 6.770 |
| Image‑Text Fusion ‑ P2 | 79.3 | 74.0 | 71.7 | 91.2 | 86.3 | 75.8 | 85.0 | 88.0 | 90.3 | 86.1 | 76.3 | 89.9 | 86.4 | 88.9 | 1.272 | 83.776 | 6.432 |
| Image‑Text Fusion ‑ P3 | 79.7 | 75.0 | 76.4 | 92.4 | 87.9 | 79.4 | 86.8 | 88.7 | 91.0 | 87.1 | 78.4 | 91.0 | 88.7 | 86.9 | 1.230 | 85.236 | 5.670 |

**T A B L E** 14 Gender Classification Accuracy (%) of Image‑Text Fusion technique on UTKFace test set across different demographics. M: Male, F: Female. Max/Min represents the ratio of maximum to minimum classification accuracy among gender and ethnicity; Overall is the average accuracy, and DoB refers to the standard deviation. T1: Text Generated using BLIP Image‑to‑Text Model; T2: Text generated using all annotated attributes; P1 to P3 are Text generated using attributes obtained from the prompts in Table 12.

| Method | Asian | | Black | | Indian | | White | | Max/Min ↓ | Overall ↑ | DoB ↓ |
|---|---|---|---|---|---|---|---|---|---|---|---|
| | M | F | M | F | M | F | M | F | | | |
| Image‑Text Fusion ‑ T1 | 84.076 | 89.189 | 96.970 | 87.727 | 96.018 | 88.890 | 95.247 | 91.087 | **1.153** | **92.080** | **4.258** |
| Image‑Text Fusion ‑ T2 | 73.885 | 93.513 | 97.403 | 85.450 | 92.478 | 87.134 | 92.505 | 86.304 | 1.318 | 89.349 | 6.760 |
| Image‑Text Fusion ‑ P1 | 66.242 | 94.595 | 96.970 | 86.360 | 91.593 | 85.380 | 91.773 | 87.174 | 1.464 | 88.712 | 8.894 |
| Image‑Text Fusion ‑ P2 | 66.242 | 93.513 | 96.970 | 85.450 | 91.150 | 87.134 | 89.762 | 86.522 | 1.464 | 87.983 | 8.660 |
| Image‑Text Fusion ‑ P3 | 94.594 | 63.694 | 96.104 | 89.090 | 91.150 | 92.400 | 89.400 | 90.435 | 1.509 | 88.911 | 9.600 |

**T A B L E** 15 Gender Classification Accuracy (%) of Image‑Text Fusion technique on DiveFace test set across different demographics. M: Male, F: Female. Max/Min is the ratio of maximum to minimum classification accuracy values among gender and ethnicity; Overall is the average accuracy, and DoB refers to the standard deviation. T1: Text Generated using BLIP Image‑to‑Text Model; T2: Text generated using all annotated attributes; P1 to P3 are Text generated using attributes obtained from the prompts in Table 12.

| Method | East Asian | | Sub‑Saharan and South Indian | | White | | Max/Min ↓ | Overall ↓ | DoB ↓ |
|---|---|---|---|---|---|---|---|---|---|
| | M | F | M | F | M | F | | | |
| Image‑Text Fusion ‑ T1 | 95.83 | 98.95 | 96.00 | 96.31 | 99.03 | 97.09 | **1.033** | **97.218** | **1.326** |
| Image‑Text Fusion ‑ T2 | 95.01 | 98.77 | 96.08 | 94.68 | 99.12 | 95.77 | 1.047 | 96.591 | 1.741 |
| Image‑Text Fusion ‑ P1 | 93.61 | 99.17 | 96.58 | 94.28 | 98.72 | 96.92 | 1.059 | 96.569 | 2.061 |
| Image‑Text Fusion ‑ P2 | 94.70 | 98.34 | 96.95 | 94.10 | 98.50 | 96.26 | 1.047 | 96.488 | 1.665 |
| Image‑Text Fusion ‑ P3 | 95.78 | 98.55 | 96.08 | 94.37 | 98.81 | 96.08 | 1.047 | 96.629 | 1.574 |

# 5 | CONCLUSION

This research pioneers the integration of textual guidance to enhance fairness in facial image‑based gender classification systems. By leverag‑ ing semantic information from captions during training, the proposed approach improves model generalization. Extensive experiments vali‑ date the efficacy of the image‑text matching and fusion methodologies in mitigating bias and enhancing accuracy across gender‑racial groups, outperforming state‑of‑the‑art methods. The introduction of textual supervision during training marks a significant step toward fostering eq‑ uitable learning, highlighting the potential of text‑guided paradigms to address algorithmic bias.

Future work will focus on extending this methodology to other bias‑prone facial analysis tasks, such as age, ethnicity, and emotion clas‑ sification. Additionally, we aim to explore the generation of optimal textual descriptions for images using large language models, examine the impact of stylistic variations in textual inputs, and incorporate multi‑ modal cues like audio. Addressing the adaptability of these systems to dynamic data distributions remains an important direction for future research. This study serves as an initial step toward interpretable and unbiased facial classification through cross‑modal learning.

## ACKNOWLEDGMENTS

The work was done when Anoop was a Ph.D. student at the School of Computing, Wichita State University. Special thanks to the advisor, Dr. Ajita Rattani, for her valuable contributions.

## CONFLICT OF INTEREST

The authors declare no potential conflict of interests.